\documentclass[journal]{IEEEtran}

\usepackage[T1]{fontenc}
\usepackage{lmodern}
\usepackage{amsmath,amssymb}
\usepackage{booktabs}
\usepackage{graphicx}
\usepackage{multirow}
\usepackage{xcolor}
\usepackage[normalem]{ulem}
\usepackage{float}

\begin{document}

\title{Class-Competence Aggregation: Output-Layer Routing for
Federated Multiclass DDoS Detection under Class-Concentrated Non-IID Data}

\author{Tapu~Datta and Hafiz~Imtiaz$^*$
\thanks{$^*$ Corresponding author.\\
T. Datta and H. Imtiaz are with the Department of Electrical and
Electronic Engineering, Bangladesh University of Engineering and
Technology (BUET), Dhaka 1205, Bangladesh (e-mail:
1024062523@eee.buet.ac.bd; hafizimtiaz@eee.buet.ac.bd).}}

\maketitle

\begin{abstract}
Federated learning over distributed intrusion-detection sensors
naturally produces \emph{class-concentrated} non-IID data, where an
entire attack class is held by a single client and absent
elsewhere. We show this collapses federated averaging in a
specific, localized way: the output layer holds one decision boundary
per class, and when a class's aggregation weight comes
overwhelmingly from clients that have never seen it, its
neuron receives an active anti-signal rather than mere dilution,
collapsing to zero. On CIC-DDoS2019, standard FedAvg drives the
concentrated LDAP class to $0.00 \pm 0.00\%$ F1 across every seed. We
propose Class-Competence Aggregation (CCA), which aggregates the
output layer per class using each client's own locally reported
validation F1 as the weight, while hidden layers remain under
standard size-weighted FedAvg. CCA requires no server-held auxiliary
data and adds only one F1 scalar per class per round beyond the feature-scaling and class-count
exchanges common to any federated pipeline. On our primary CNN-LSTM
architecture, CCA alone raises LDAP F1 from $0\%$ to $84.83 \pm 1.65\%$,
and the full method reaches $91.64 \pm 0.57\%$ macro F1,
closing the federated-to-centralized gap from $-42.53$ to $-4.05$
points. Benchmarked against four other federated methods under
identical partitions and seeds, CCA is the only one that recovers the
concentrated class substantially and reliably; a $178\times$ smaller
MLP shows the same qualitative recovery, though less completely,
confirming the effect holds across architecture scale even as its
magnitude varies. Results are validated across three
seeds and replicated on a second dataset, \mbox{CICIDS2017}.
\end{abstract}


\section{Introduction}
\label{sec:intro}

Network intrusion detection in the real world is a distributed task:
privacy regulation and inter-organizational confidentiality often
prohibit pooling raw flow records across organizations, and streaming
them to a central collector is bandwidth-prohibitive in any case.
Detection instead runs at each organization's own vantage
point: edge routers, ingress points, cloud gateways, individual
service endpoints. Therefore, no single sensor sees the full traffic mix.
Federated learning \cite{fedavg2017} is a natural fit here: it trains
a shared model from these local vantage points without requiring any
of them to give up their raw data.

In our setting, a federated binary classifier (attack vs.\ benign)
reaches $99.56\%$ macro F1, within $0.28$ points of its centralized
ceiling. This is consistent with results reported elsewhere on this
dataset: Akgun et al.\ \cite{akgun2022} report $99.99\%$ accuracy for binary
classification and $99.30\%$ for multiclass classification using a
centralized CNN-based model on CIC-DDoS2019, indicating that our own
centralized reference is in line with the literature before the
federated setting is introduced. Every client holds both benign and
attack traffic in the binary case; therefore, averaging local models yields a
near-optimal global model with no data sharing at all.

The multiclass setting breaks this. FedAvg does not merely lose some
accuracy under class-concentrated non-IID data; it destroys the
learned representation for the concentrated class in the output layer
specifically. A sensor at an LDAP service captures LDAP amplification
attacks; a sensor at an MSSQL endpoint captures MSSQL amplification;
sensors on high-volume paths capture Syn and UDP floods. This
partitioning of \emph{attack visibility} produces what we call the
class-concentrated non-IID regime, in which LDAP and
MSSQL are concentrated almost exclusively (over $98\%$) at a single client
each, absent from all others but a trivial tail; Syn and UDP are primarily
associated with one client but retain a substantial secondary presence at
the mixed \texttt{Client\_UDPLag} (Table~\ref{tab:partition}).
Zhao et al.\ \cite{zhao2018} identified this as the most severe form of federated
data heterogeneity, but their remedy, maintaining a small globally
shared dataset, reintroduces precisely the data sharing that federated
learning exists to avoid.

The failure has a precise mechanism, not just a low score. Every
client trains the same five-class architecture; therefore, even a client that
has never seen a single LDAP sample still produces an ``LDAP neuron''
after local training: the question is what that neuron encodes. At
\texttt{Client\_LDAP}, real LDAP examples push the neuron's weights
toward correctly firing on LDAP traffic. But cross-entropy loss does
not act on one class in isolation: rewarding the correct class on a
training step simultaneously pushes down the logits of every other
class. This means that at the clients holding no LDAP data, ordinary
training on their own Syn, UDP, MSSQL, and Benign traffic actively
pushes their LDAP-neuron weights in the \emph{opposite}
direction: toward suppressing LDAP activation. These clients are not
contributing noise around zero; they are contributing an opposing
signal, and because they vastly outnumber the one specialist in
aggregation weight, standard FedAvg's dataset-size-weighted average is
dominated by suppression rather than merely diluted. We work out this
imbalance in exact figures once the partition is introduced
(Section~\ref{sec:setup}, Table~\ref{tab:partition}) and formalize the
mechanism in Section~\ref{sec:failure}. \texttt{Client\_LDAP} must relearn
its specialization from scratch each round, only to have it
overwritten again. In our five-client partition of CIC-DDoS2019
\cite{sharafaldin2019, dhoogla-cicddos2019}, this produces total failure: LDAP
F1 $= 0.00 \pm 0.00\%$ across every seed, with macro F1 falling
$42.53$ points below centralized training.

Our central observation is that this failure is \emph{not} uniformly
distributed across the network: it is confined to the output layer,
the one place where a parameter block belongs to a single class
(Section~\ref{sec:failure} formalizes why hidden layers are unaffected).

This suggests a correction that is targeted rather than global. We
introduce \textbf{Class-Competence Aggregation (CCA)}: after each
round, every client reports its per-class validation F1, and the server
uses these as aggregation weights for the output layer only. A client
with no examples of class $c$ reports $\text{F1}_c \approx 0$ and
receives near-zero weight for that class's neuron; the specialist
client receives near-total weight. Hidden layers remain under standard
size-weighted FedAvg. No server-held data and no inter-client sharing
are required.

Our contributions are:

\begin{enumerate}
  \item We propose CCA, a federated aggregation mechanism that applies
    per-class F1-weighted averaging exclusively to the output layer.
    Added alone to a class-weighted, adaptively scheduled federated
    pipeline, it raises CNN-LSTM LDAP F1 from $0.00 \pm 0.00\%$ to
    $84.83 \pm 1.65\%$ and macro F1 by $20.00$ points: the largest and
    most reproducible single effect in our ablation.
  \item We show empirically that the output layer is the bottleneck.
    Because CCA corrects aggregation nowhere else, its effect isolates
    the locus of failure. Changing how a single layer is averaged, out
    of $1.29$M parameters in the CNN-LSTM, takes LDAP F1 from total
    collapse ($0.00\%$) to within $0.5$ points of the centralized
    ceiling ($86.66\%$): the best achievable with no federation
    constraints at all.
  \item We report a reliability result. Benchmarked against FedAvg,
    FedProx, FedNova, and FedDWC under identical partitions, splits and
    seeds, CCA is the only method that recovers the concentrated class
    both substantially and reliably. FedAvg and FedProx fail completely
    and consistently ($0.00 \pm 0.00\%$); FedNova and FedDWC recover it
    only on some seeds, with standard deviations of $40.8$ and $23.5$
    points. CCA's is under $1$ point.
  \item We show that single-run ablations can misattribute credit. A
    single unseeded run of one ablation configuration reported
    $85.65\%$ LDAP F1; the same configuration under three-seed
    validation yields $0.00 \pm 0.00\%$. We report the seeded results
    throughout and treat this as a methodological finding in its own
    right.
\end{enumerate}

\section{Related Work}
\label{sec:related}

\subsection{Non-IID federated learning}

McMahan et al.\ \cite{fedavg2017} introduced FedAvg, aggregating client weights in
proportion to local dataset size, and reported robustness to non-IID
data (on a 2-class-per-client partition). Zhao et al.\ \cite{zhao2018} showed this
robustness degrades sharply in the 1-class regime, with CIFAR-10
accuracy falling $51.31\%$, and attributed the loss to weight
divergence between locally homogeneous clients. Their proposed fix is a
globally shared dataset, unavailable under our privacy constraint.
Li et al.\ \cite{fedprox2020} proposed FedProx, adding a proximal penalty
$\frac{\mu}{2}\|w_{\text{local}} - w_{\text{global}}\|^2$ to constrain
client drift. FedProx targets quantity skew, where clients differ in
volume and balance but each still holds most classes; we show its
behavior under class concentration is architecture-dependent
(Section~\ref{sec:results}). FedNova \cite{fednova2020} normalizes for
unequal local computation, a distinct axis of heterogeneity. A broader
family targets general non-IID optimization without addressing class
concentration specifically: SCAFFOLD \cite{scaffold2020} corrects
client drift with control variates but still averages every layer
uniformly; MOON \cite{moon2021} adds a model-contrastive loss between
local and global representations; FedBN \cite{fedbn2021} excludes
batch-norm parameters from aggregation entirely, addressing
feature-distribution shift rather than a class being absent from all
but one client; clustered FL \cite{sattler2021clustered} trains a
separate model per client cluster, trading away the cross-client
generalization a single global model provides on shared classes;
adaptive server optimizers \cite{reddi2021} improve convergence
orthogonally to which parameters are aggregated how. Kairouz et al.\ \cite{kairouz2021}
survey this landscape and list class imbalance among federated
learning's open problems without a solution specific to output-layer
concentration. Separately, Wang et al.\ \cite{wang2021ratio} address general
(non-IDS) class imbalance with Ratio Loss, a training-level reweighting
similar in spirit to our class-weighted loss (Section~\ref{sec:preconditions})
that, like ours alone, does not correct aggregation-level dilution.

\subsection{Federated intrusion detection}

Federated IDS was established as viable by Rahman et al.\ \cite{rahman2020iotids} and
Nguyen et al.\ \cite{nguyen2019diot}'s D\"{I}oT, neither addressing multiclass
non-IID structure; Agrawal et al.\ \cite{agrawal2022survey} survey the field and flag
statistical heterogeneity as open, consistent with the gap below.
FLAD \cite{flad2023} is the closest prior work: it schedules local
epochs adaptively, giving more training time to clients the global
model explains worst. FLAD was designed and evaluated for
\emph{binary} DDoS detection and aggregates by unweighted arithmetic
mean, which cannot correct output-neuron dilution. FedMADE
\cite{fedmade2024} computes a class probability matrix per client
using a server-held auxiliary set (10 samples per class) and solves for
cluster weights; FedDWC \cite{feddwc2025} adjusts per-client weights
by round-over-round accuracy change. In both cases, each client
receives a single scalar competence score, and that one score scales
\emph{every parameter} in that client's model: hidden layers and
output layer alike, with no per-class or per-layer distinction.
FedDWC's signal is additionally
counterproductive here: overall accuracy correlates with data volume;
therefore, the large clients are rewarded and \texttt{Client\_LDAP}, the sole
holder of LDAP knowledge, is penalized, amplifying the imbalance that
already causes the failure. DS-FedIDS \cite{dsfedids2025} rebalances
each client's local dataset by resampling, addressing intra-client
imbalance; resampling cannot create signal at a client with zero
examples of a class; therefore, it is complementary rather than competing.

\subsection{Gap}

Table~\ref{tab:related} summarizes. No prior FL-IDS method (i)
constructs or evaluates a class-concentrated partition in which an
attack class is held entirely by one sensor, or (ii) restricts
performance-based aggregation to the output layer, where the
class-concentrated failure is structurally localized.

\begin{table*}[t]
\centering
\caption{Positioning against the most closely related federated intrusion detection methods.}
\label{tab:related}
\vspace{4pt}
\small
\resizebox{\linewidth}{!}{%
\begin{tabular}{lccccc}
\toprule
\textbf{Dimension} & \textbf{FLAD} & \textbf{FedMADE} & \textbf{FedDWC} & \textbf{DS-FedIDS} & \textbf{This work} \\
\midrule
Non-IID type       & ---         & General het.   & General het. & Intra-client & Class-concentrated \\
Task               & Binary DDoS & IDS (7-class)  & DDoS IDS     & IDS (multi)  & 5-class DDoS \\
Aggregation scope  & Full model  & Full model     & Full model   & Full model   & Output layer only \\
Weight signal      & ---         & Cluster perf.  & Round acc.   & ---          & Per-class F1 \\
Server aux.\ data  & No          & Yes (10/class) & No           & No           & No \\
Multi-seed validated & ---       & ---            & ---          & ---          & 3 seeds \\
\bottomrule
\end{tabular}%
}
\end{table*}

\section{Method}
\label{sec:method}

\subsection{Why FedAvg Collapses the Concentrated Class}
\label{sec:failure}

FedAvg aggregates client parameters proportionally to local dataset
size:
\begin{equation}
  w_{\text{global}} = \sum_{i=1}^{K} \frac{n_i}{n_{\text{total}}}\, w_i .
  \label{eq:fedavg}
\end{equation}
For the final linear layer, class logits are computed as
\begin{equation}
  \text{logit}_c = W_{\text{out}}[c,:] \cdot h + b_c ,
  \label{eq:logit}
\end{equation}
where $h$ is the penultimate representation. The row
$W_{\text{out}}[c,:]$ together with $b_c$ constitutes a dedicated
linear decision boundary for class $c$; we refer to this pair as the
\emph{output neuron} for $c$.

Consider the LDAP output neuron after one round of local training.
\texttt{Client\_LDAP} has received gradient updates from $1{,}884$ LDAP
samples, pushing $W_{\text{out}}[c_{\text{LDAP}},:]$ toward the
direction in feature space that discriminates LDAP from all other
classes. Every other client has observed only Benign, Syn, UDP, or
MSSQL traffic. For those clients, LDAP is a class that never appears as
the target label; therefore, cross-entropy pushes
$W_{\text{out}}[c_{\text{LDAP}},:]$ in directions that \emph{suppress}
LDAP activation on the traffic they do see. Their contribution is not
noise around zero; it is systematic anti-signal.

Eq.~\ref{eq:fedavg} then combines these five vectors with the
size-proportional weights of Table~\ref{tab:partition},
$[0.108,\, 0.183,\, 0.250,\, 0.250,\, 0.210]$. The specialist's
contribution is outvoted roughly eight to one, the aggregated neuron is
destroyed, and \texttt{Client\_LDAP} must rebuild it from scratch in
the following round. This loop never terminates in a useful state:
LDAP F1 is $0.00\%$ at every round of every seed.

Two properties of this analysis matter for the design that follows.
First, the failure is confined to the output layer, because only there
is a parameter block dedicated to a single class. Second, the fix
requires no data: the server needs only to know \emph{which client is
competent for which class}, a quantity each client can compute locally.

\subsection{Class-Competence Aggregation}
\label{sec:cca}

\begin{figure*}[t]
\centering
\includegraphics[width=\linewidth]{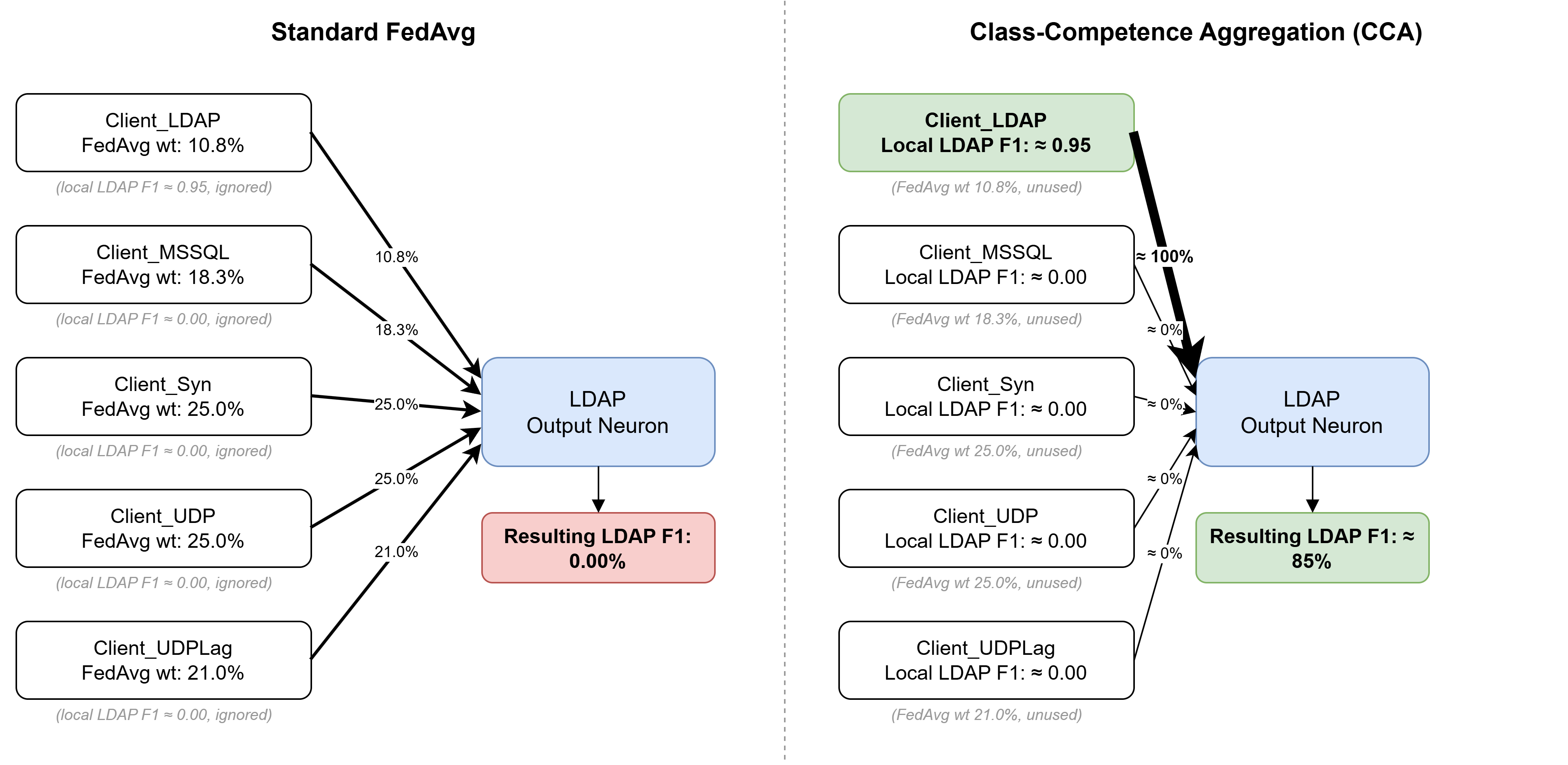}
\caption{Standard FedAvg (left) vs.\ CCA (right) for the LDAP class.
Line thickness represents aggregation weight. Under FedAvg, all five
clients contribute comparably regardless of whether they hold any
LDAP data. Under CCA, weight for the LDAP output row is routed almost
entirely to \texttt{Client\_LDAP}, the only client with meaningful
local LDAP competence.}
\label{fig:cca}
\end{figure*}

Figure~\ref{fig:cca} contrasts the two aggregation rules side by side.
Standard FedAvg (left) weights every client by dataset size alone;
therefore, all five clients contribute comparably to the hidden layers. CCA
(right) instead weights each client's contribution to a given class's
output row by that client's own local competence on that class,
collapsing nearly all of the weight onto the specialist.

After local training in round $t$, each client $i$ evaluates its local
model on its own held-out validation split and reports per-class F1
scores $f1_{i,c}$. No data leaves the client; the reported quantity is
a scalar per class. The $f1_{i,c}$ score is computed with
scikit-learn's \texttt{zero\_division=0} convention: if client $i$'s
local validation split contains no examples of class $c$, precision
and recall are both the mathematically undefined $0/0$, and this
convention resolves them to exactly $0$ rather than an undefined
value or a warning. This is the mechanism that produces the
$f1_{i,c} \approx 0$ behavior described above, and it is applied
identically to the official-test-set per-class F1 reported throughout
Section~\ref{sec:results}.

Raw per-round F1 is noisy; therefore, we smooth with an exponential moving
average:
\begin{equation}
  \widetilde{f1}_{i,c,t} = \alpha\, f1_{i,c,t} + (1-\alpha)\, \widetilde{f1}_{i,c,t-1},
  \label{eq:ema}
\end{equation}
with $\alpha = 0.7$ and $\widetilde{f1}_{i,c,0} = 0$. Weights are then
normalized across clients, per class. Rather than a
smoothed denominator, this normalization is implemented as an exact
conditional on whether any client has learned the class yet:
\begin{equation}
  \tilde{w}_{i,c} =
  \begin{cases}
    \dfrac{\widetilde{f1}_{i,c,t}}{\sum_j \widetilde{f1}_{j,c,t}} & \text{if } \sum_j \widetilde{f1}_{j,c,t} > 0, \\[4pt]
    1/K & \text{otherwise.}
  \end{cases}
  \label{eq:ccaweight}
\end{equation}
The second case, uniform weighting, applies exactly
when every client reports zero F1 for class $c$, which occurs in early
rounds before any client has learned a minority class; no smoothing
constant $\epsilon$ is added in either branch.

\paragraph{Output layer}
For output weight matrix $W_{\text{out}} \in \mathbb{R}^{C \times H}$
and bias $b_{\text{out}} \in \mathbb{R}^{C}$, each class row is
aggregated with its own weight vector:
\begin{align}
  W^{\text{global}}_{\text{out}}[c,:] &= \sum_i \tilde{w}_{i,c}\, W^i_{\text{out}}[c,:] , \label{eq:ccaW}\\
  b^{\text{global}}_{\text{out}}[c]   &= \sum_i \tilde{w}_{i,c}\, b^i_{\text{out}}[c] . \label{eq:ccab}
\end{align}

\paragraph{All other layers}
Every remaining parameter uses unmodified size-weighted FedAvg
(Eq.~\ref{eq:fedavg}). This is deliberate. Hidden representations are
built from traffic that all clients observe, and reweighting them by
per-class competence would discard genuine contributions from clients
that happen to be weak on one class while distorting the shared feature
space that the corrected output layer depends on.

The mechanism is self-selecting. Once \texttt{Client\_LDAP} reaches a
local LDAP F1 near $1.0$ and all other clients remain near $0$,
Eq.~\ref{eq:ccaweight} collapses onto the specialist, assigning it
near-total weight for that one neuron while leaving the other four
neurons and every hidden layer untouched.

\subsection{Preconditions: Adaptive Scheduling and Class Weighting}
\label{sec:preconditions}

CCA is a routing mechanism: it can only route toward competence that
already exists. Two standard components establish it.

\paragraph{Adaptive epoch scheduling}
Following FLAD \cite{flad2023}, we allocate local epochs by how poorly
the current global model explains each client's data. We use validation
loss $\ell_i$ as the scheduling signal rather than F1, since loss
remains informative under majority-class dominance in the multiclass
setting:
\begin{align}
  \tilde{\ell}_i &= \frac{\ell_i - \ell_{\min}}{\ell_{\max} - \ell_{\min} + \epsilon} , \label{eq:fladnorm}\\
  E_i &= \text{round}\!\left(E_{\min} + \tilde{\ell}_i (E_{\max} - E_{\min})\right) . \label{eq:fladep}
\end{align}
The allocation converges by round 5: harder clients (MSSQL, LDAP)
receive 7--10 local epochs, versus 1--3 for the high-volume
clients, concentrating budget exactly where FLAD intends.

\paragraph{Class-weighted loss}
Adaptive scheduling corrects imbalance \emph{between} clients but not
\emph{within} them; benign traffic is the local majority everywhere. We
apply inverse-frequency weighting to the local cross-entropy,
$\text{weight}_c = n_{\text{total}} / (|\mathcal{C}| \cdot n_c)$, which
gives each LDAP sample at \texttt{Client\_LDAP} roughly $10\times$ the
gradient contribution of a benign sample. The $n_c$ is
computed once, during offline dataset preparation, from label counts
pooled across the entire federation, not per client; since all five
retained classes have a nonzero count somewhere in the federation by
construction (Section~\ref{sec:setup}'s exclusion criteria already
rule out classes too sparse for this), $n_c = 0$ cannot occur for any
class this formula is applied to. Pooling raw label counts this way is
subject to the same simulation-vs-deployment caveat as the feature
scaler above: a live deployment could instead have each client report
only its own local per-class counts (aggregate volumes, not raw
records) for the server to sum via secure aggregation, which is exact
because summation, like the min/max used for scaling, is associative
across clients.

Section~\ref{sec:ablation} shows these two components are necessary but not
sufficient: together they add $18.30$ points of macro F1 while leaving
LDAP F1 at exactly $0.00\%$.

\subsection{Implementation Details}

We add FedProx regularization \cite{fedprox2020} with $\mu = 0.01$ and
track the best-validation-macro-F1 checkpoint rather than taking the
final round, since CCA's round-to-round weight updates make training
non-monotonic. Learning rate decays as
$\text{lr}_t = 10^{-3} \times 0.95^{(t-1)}$. Both are stabilizers, not
sources of the effect; their measured contributions are reported in
Section~\ref{sec:ablation}.

\section{Experimental Setup}
\label{sec:setup}

\paragraph{Data and partition}
We use the Kaggle Parquet redistribution of CIC-DDoS2019
\cite{sharafaldin2019, dhoogla-cicddos2019}, with 77 CICFlowMeter
features per flow. Of the
twelve attack categories, four (DNS, NTP, SNMP, TFTP) are absent from
the training partition, and three (NetBIOS, Portmap, UDPLag) have fewer
than ten official test samples each, making per-class F1 statistically
meaningless. We retain five classes: Benign, LDAP, MSSQL, Syn, UDP. The
four attack classes span both mechanistic families of the dataset,
reflection-based (LDAP, MSSQL) and exploitation-based (Syn, UDP).

Five clients are constructed by assigning four of them the traffic for
one primary attack type each; the fifth, \texttt{Client\_UDPLag},
receives the residual Syn and UDP traffic left over from that
assignment and therefore holds a mixed attack profile rather than a single
specialization (Table~\ref{tab:partition}). Every client also holds a
benign share; therefore, the concentration is of \emph{attack} classes
specifically. LDAP is $98.8\%$ concentrated at one client.
Clients whose raw data exceeds a fixed cap of
$15{,}000$ samples are subsampled once, with a fixed seed, before the
three federated training seeds: within each such client, every class
is sampled in proportion to its own share of that client's data;
therefore, the client's internal class mixture, the property this partition is
designed around, is preserved rather than disturbed by the cap.
Integer truncation of each class's proportional allocation means the
resulting per-client totals land just under $15{,}000$ rather than
exactly at it, which is why Table~\ref{tab:partition} reports
$14{,}999$ for \texttt{Client\_Syn} and \texttt{Client\_UDP}.

\begin{table*}[t]
\centering
\caption{Full class-by-client sample count matrix for the training
partition. LDAP appears at two clients ($1{,}884$ at
\texttt{Client\_LDAP} + $22$ at \texttt{Client\_MSSQL}); MSSQL likewise
at two ($8{,}378$ + $122$ at \texttt{Client\_UDP}); Syn and UDP are each
split between their primary client and the mixed \texttt{Client\_UDPLag}.
Benign traffic (column 1) is present at every client.}
\label{tab:partition}
\vspace{4pt}
\small
\setlength{\tabcolsep}{4pt}
\begin{tabular}{lrrrrrrr}
\toprule
\textbf{Client} & \textbf{Benign} & \textbf{LDAP} & \textbf{MSSQL} & \textbf{Syn} & \textbf{UDP} & \textbf{Total} & \textbf{FedAvg wt.} \\
\midrule
Client\_LDAP   & 4,585 & 1,884 &     0 &     0 &      0 &  6,469 & 10.8\% \\
Client\_MSSQL  & 2,574 &    22 & 8,378 &     0 &      0 & 10,974 & 18.3\% \\
Client\_Syn    & 5,765 &     0 &     0 & 9,234 &      0 & 14,999 & 25.0\% \\
Client\_UDP    & 2,391 &     0 &   122 &     0 & 12,486 & 14,999 & 25.0\% \\
Client\_UDPLag & 3,748 &     0 &     0 & 5,538 &  3,298 & 12,584 & 21.0\% \\
\midrule
Total          & 19,063 & 1,906 & 8,500 & 14,772 & 15,784 & 60,025 & 100.0\% \\
\bottomrule
\end{tabular}
\end{table*}

\paragraph{Evaluation}
Each client splits locally into $85\%$ train / $15\%$ validation. All
models, federated and centralized, are evaluated on the \emph{official}
CIC-DDoS2019 test partition ($70{,}009$ samples; Benign $73.4\%$, UDP
$14.9\%$, MSSQL $8.9\%$, LDAP $2.1\%$, Syn $0.8\%$). A global MinMax
scaler fitted on pooled federated training data is applied without
refitting; therefore, no test statistics leak. This scaler was fitted
once during offline preparation of the CIC-DDoS2019 benchmark, before
any client--server protocol runs, since our federation is simulated by
partitioning a dataset we already hold in full. A live deployment could
reproduce the same global scaler without pooling raw data: each client
reports only its local per-feature minimum and maximum (two scalars
per feature) in a one-time setup exchange before training begins, and
the server merges these into the exact global min/max (min and max are
associative statistics; therefore, this is mathematically identical to fitting
on pooled data). Unlike CCA's per-round F1 reporting
(Section~\ref{sec:cca}), this exchange happens only once, and each
disclosed value is a genuine observation from the client's data rather
than a derived aggregate. We report macro F1: with benign
at $73.4\%$ of the test set, accuracy would mask complete failure on
the minority classes.

\begin{figure}[t]
\centering
\includegraphics[width=\linewidth]{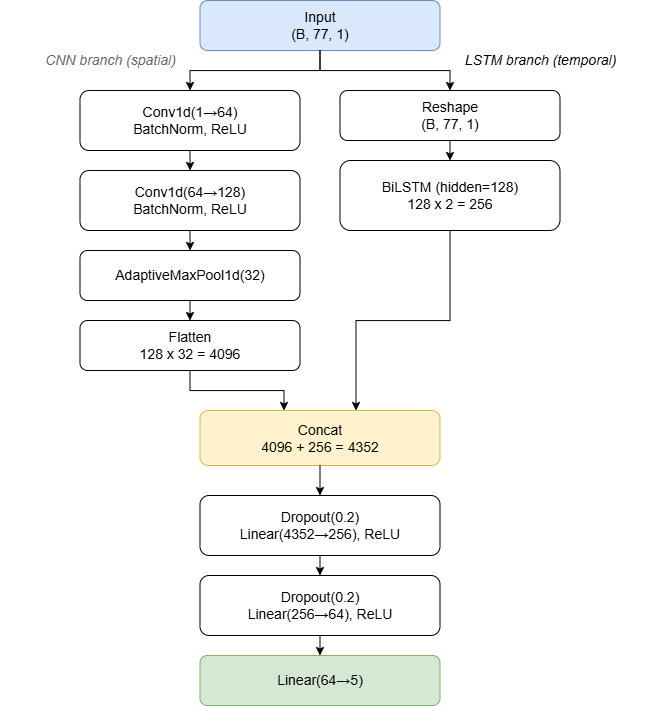}
\caption{The dual-branch CNN-LSTM architecture ($1{,}290{,}629$
parameters). The two branches are fused before the final 5-way output
layer, whose row-wise structure (Eq.~\ref{eq:logit}) is what CCA
targets.}
\label{fig:architecture}
\end{figure}

\paragraph{Architectures}
Figure~\ref{fig:architecture} shows the primary model, a dual-branch
CNN-LSTM ($1{,}290{,}629$ parameters). We also evaluate a three-layer MLP ($178\times$ smaller):
$\text{Linear}(77{\to}64) \to \text{ReLU} \to \text{Linear}(64{\to}32)
\to \text{ReLU} \to \text{Linear}(32{\to}5)$, $7{,}237$ parameters
total, to test whether the findings are architecture-dependent.

\paragraph{Reproducibility}
Every reported federated configuration is run under three fixed seeds
(42, 123, 7), with PyTorch, CUDA, NumPy and Python RNGs seeded and
cuDNN set deterministic before any data loading.All models are implemented in PyTorch 2.10.0 (CUDA 12.8) and optimized with Adam, batch size 64, for a fixed 30 federated rounds; training data is reshuffled each local epoch, and all layers use PyTorch's default initialization (no custom scheme applied). We report
mean $\pm$ standard deviation. Training used a T4×2 GPU.
Significance is assessed with a paired $t$-test over the three seeds;
we note in Section~\ref{sec:limitations} that $n{=}3$ has severely limited
power. Code and experimental configurations will be made available upon publication.

\section{Results}
\label{sec:results}

\subsection{The Gap}

Centralized CNN-LSTM reaches $95.69\%$ macro F1 with LDAP at $86.66\%$;
LDAP is the hardest class even with full data access. Federated
training with standard FedAvg reaches $53.16 \pm 8.29\%$ macro F1 with
LDAP at exactly $0.00 \pm 0.00\%$. Benign
($99.35 \pm 0.89\%$) is near-perfect and stable across seeds, while
Syn's mean of $76.71 \pm 32.41\%$ conceals large per-seed swings
rather than consistently near-perfect performance: the global model
has learned the majority class reliably, is unreliable on Syn, and
fails entirely on the concentrated class. This is the
$-42.53$-point gap the method aims to close.

\subsection{Comparison Against Federated Baselines}

We re-implement FedProx, FedNova and FedDWC in their canonical
published form (fixed local epochs, plain cross-entropy, no adaptive
scheduling) and run them under the identical partition, splits, test
set and seeds. Table~\ref{tab:baselines} reports CNN-LSTM.

\begin{table*}[t]
\centering
\caption{CNN-LSTM, full per-class F1, mean $\pm$ std over seeds
42/123/7. Gap is macro F1 minus the centralized reference ($95.69\%$).}
\label{tab:baselines}
\vspace{4pt}
\footnotesize
\resizebox{\linewidth}{!}{%
\begin{tabular}{lccccccc}
\toprule
\textbf{Method} & \textbf{Benign} & \textbf{LDAP} & \textbf{MSSQL} & \textbf{Syn} & \textbf{UDP} & \textbf{Macro F1} & \textbf{Gap} \\
\midrule
FedAvg   & $99.35 \pm 0.89$ & $0.00 \pm 0.00$   & $10.02 \pm 13.83$ & $76.71 \pm 32.41$ & $79.73 \pm 0.97$ & $53.16 \pm 8.29$ & $-42.53$ \\
FedProx  & $99.91 \pm 0.02$ & $0.00 \pm 0.00$   & $20.12 \pm 4.69$  & $95.01 \pm 1.07$  & $79.44 \pm 0.68$ & $58.90 \pm 1.22$ & $-36.79$ \\
FedNova  & $99.96 \pm 0.02$ & $28.84 \pm 40.79$ & $14.24 \pm 9.10$  & $97.60 \pm 1.96$  & $79.06 \pm 0.66$ & $63.94 \pm 6.48$ & $-31.75$ \\
FedDWC   & $99.66 \pm 0.23$ & $66.39 \pm 23.45$ & $29.39 \pm 21.94$ & $39.84 \pm 42.64$ & $55.14 \pm 39.10$ & $58.09 \pm 9.75$ & $-37.60$ \\
\textbf{CCA (ours)} & $\mathbf{99.88 \pm 0.01}$ & $\mathbf{86.69 \pm 0.81}$ & $\mathbf{83.84 \pm 0.99}$ & $\mathbf{94.63 \pm 1.12}$ & $\mathbf{93.13 \pm 0.54}$ & $\mathbf{91.64 \pm 0.57}$ & $\mathbf{-4.05}$ \\
\bottomrule
\end{tabular}%
}
\end{table*}

\begin{table*}[t]
\centering
\caption{MLP, full per-class F1, mean $\pm$ std over seeds 42/123/7.
Gap is macro F1 minus the centralized reference ($94.66\%$).}
\label{tab:baselinesmlp}
\vspace{4pt}
\footnotesize
\resizebox{\linewidth}{!}{%
\begin{tabular}{lccccccc}
\toprule
\textbf{Method} & \textbf{Benign} & \textbf{LDAP} & \textbf{MSSQL} & \textbf{Syn} & \textbf{UDP} & \textbf{Macro F1} & \textbf{Gap} \\
\midrule
FedAvg   & $99.63 \pm 0.07$ & $0.00 \pm 0.00$ & $61.76 \pm 11.75$ & $28.68 \pm 15.95$ & $87.97 \pm 2.99$ & $55.61 \pm 6.04$ & $-39.05$ \\
FedProx  & $99.84 \pm 0.06$ & $0.00 \pm 0.00$ & $71.40 \pm 1.29$  & $93.52 \pm 1.75$  & $90.22 \pm 0.69$ & $71.00 \pm 0.74$ & $-23.66$ \\
FedNova  & $99.57 \pm 0.09$ & $0.17 \pm 0.24$  & $38.27 \pm 31.72$ & $26.03 \pm 27.63$ & $85.61 \pm 5.10$ & $49.93 \pm 6.61$ & $-44.73$ \\
FedDWC   & $93.94 \pm 3.75$ & $0.37 \pm 0.19$  & $0.00 \pm 0.00$   & $65.27 \pm 1.48$  & $0.28 \pm 0.04$  & $31.97 \pm 0.75$ & $-62.69$ \\
\textbf{CCA (ours)} & $\mathbf{99.86 \pm 0.00}$ & $\mathbf{85.40 \pm 2.27}$ & $\mathbf{78.69 \pm 1.99}$ & $\mathbf{92.67 \pm 0.31}$ & $\mathbf{91.34 \pm 0.47}$ & $\mathbf{89.59 \pm 0.90}$ & $\mathbf{-5.07}$ \\
\bottomrule
\end{tabular}%
}
\end{table*}

Two failure modes appear. FedAvg and FedProx fail on the concentrated
class \emph{completely and consistently}: neither corrects output-layer
dilution; therefore, LDAP F1 is exactly zero in every seed. FedNova and FedDWC
\emph{sometimes} recover it, but with standard deviations of $40.79$ and
$23.45$ points, recovering strongly on one seed and collapsing on
others. Neither targets the output layer; therefore, when they succeed it is an
incidental side effect of broader reweighting rather than a mechanism.

We consider this reliability gap as informative as the mean: a
standard deviation under one point, against $23$--$41$ points for the
two methods that occasionally succeed, suggests targeted output-layer
correction is a different \emph{kind} of intervention, not a
better-tuned instance of general-purpose reweighting. The ordering
holds for the MLP too (Table~\ref{tab:baselinesmlp}): FedDWC degrades to $31.97 \pm 0.75\%$, below the
FedAvg baseline, while our method reaches $89.59 \pm 0.90\%$.

\subsection{Ablation}
\label{sec:ablation}

Table~\ref{tab:ablation} decomposes the method in dependency order:
training-level components first (they must be active before CCA has
competence to route), then the aggregation-level fix, then stabilizers.

\begin{table*}[t]
\centering
\caption{CNN-LSTM ablation, full per-class F1, three seeds each. CW = class-weighted loss.
The FLAD+CW $\rightarrow$ +CCA transition is significant at $p = 0.006$
(paired $t$-test, $n{=}3$); proposed vs.\ FedAvg at $p = 0.020$.}
\label{tab:ablation}
\vspace{4pt}
\footnotesize
\resizebox{\linewidth}{!}{%
\begin{tabular}{lccccccc}
\toprule
\textbf{Configuration} & \textbf{Benign} & \textbf{LDAP} & \textbf{MSSQL} & \textbf{Syn} & \textbf{UDP} & \textbf{Macro F1} & \textbf{Gap} \\
\midrule
Centralized (reference)     & 99.97 & 86.66 & 94.09 & 99.44 & 98.30 & 95.69 & --- \\
\midrule
FedAvg                      & $99.35 \pm 0.89$ & $0.00 \pm 0.00$ & $10.02 \pm 13.83$ & $76.71 \pm 32.41$ & $79.73 \pm 0.97$ & $53.16 \pm 8.29$ & $-42.53$ \\
FLAD only                   & $99.96 \pm 0.01$ & $0.00 \pm 0.00$ & $44.31 \pm 14.79$ & $98.02 \pm 1.22$  & $84.07 \pm 3.37$ & $65.27 \pm 3.71$ & $-30.42$ \\
FLAD + CW                   & $99.93 \pm 0.03$ & $0.00 \pm 0.00$ & $70.74 \pm 13.56$ & $95.23 \pm 3.12$  & $91.41 \pm 4.34$ & $71.46 \pm 3.25$ & $-24.23$ \\
FLAD + CW + \textbf{CCA}    & $99.89 \pm 0.06$ & $\mathbf{84.83 \pm 1.65}$ & $86.36 \pm 2.31$ & $91.40 \pm 6.33$ & $94.83 \pm 1.08$ & $\mathbf{91.46 \pm 1.16}$ & $-4.23$ \\
\midrule
\ \ + FedProx               & $99.83 \pm 0.02$ & $86.05 \pm 1.22$ & $82.67 \pm 0.55$ & $94.00 \pm 0.07$ & $92.40 \pm 0.11$ & $90.99 \pm 0.32$ & $-4.70$ \\
\ \ + FedProx + best ckpt   & $99.83 \pm 0.01$ & $86.65 \pm 1.15$ & $83.13 \pm 0.45$ & $94.82 \pm 1.10$ & $92.55 \pm 0.12$ & $91.40 \pm 0.51$ & $-4.29$ \\
\ \ + all (proposed)        & $99.88 \pm 0.01$ & $86.69 \pm 0.81$ & $83.84 \pm 0.99$ & $94.63 \pm 1.12$ & $93.13 \pm 0.54$ & $91.64 \pm 0.57$ & $-4.05$ \\
\bottomrule
\end{tabular}%
}
\end{table*}

\begin{figure}[H]
\centering
\includegraphics[width=\linewidth]{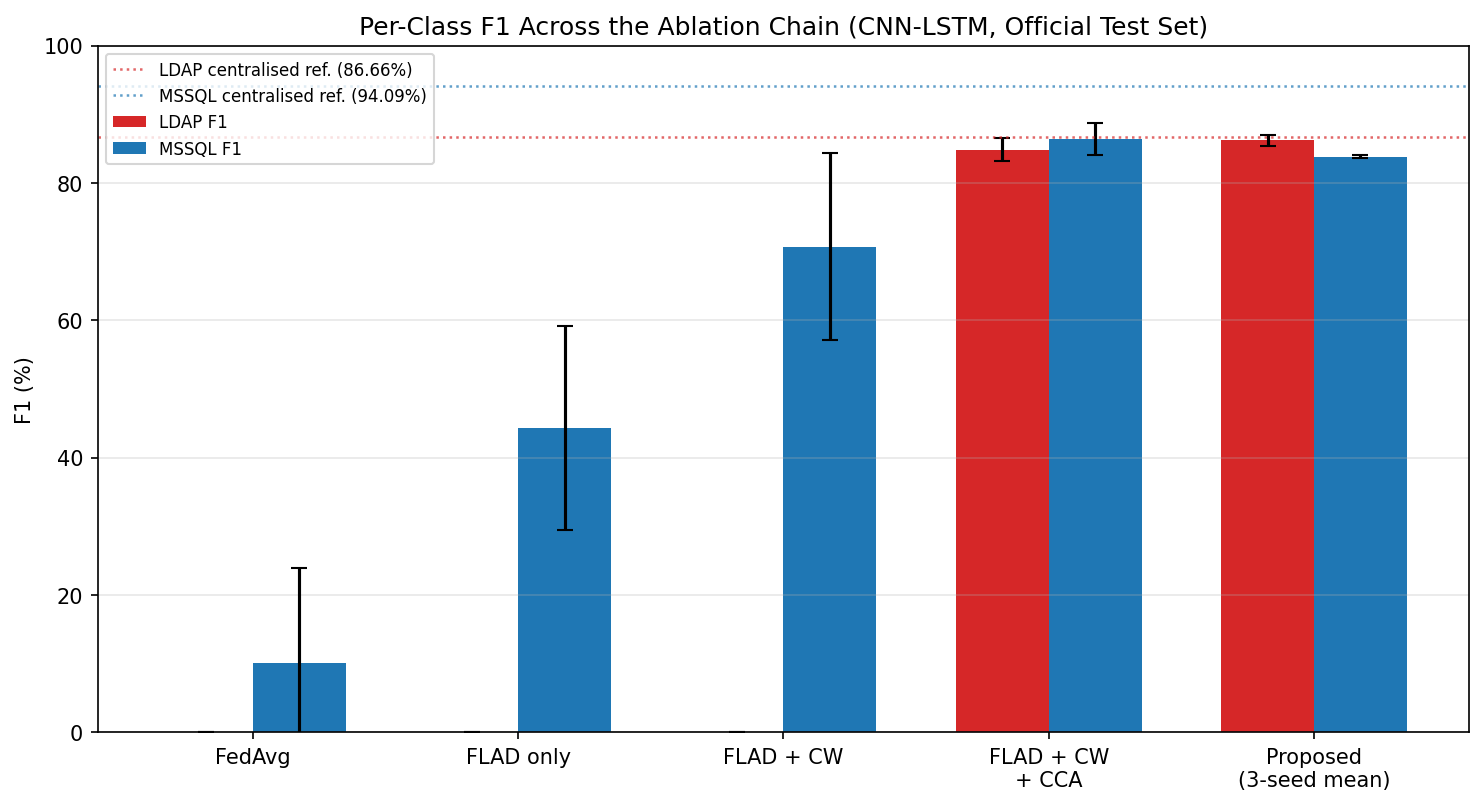}
\caption{LDAP and MSSQL F1 across the ablation chain (mean $\pm$ std
over seeds 42/123/7 where seeded). LDAP F1 stays at exactly $0.00\%$
through FedAvg, FLAD-only, and FLAD+CW, and only recovers once CCA is
added. The dotted lines mark the centralized (non-federated) per-class
F1 ceiling for LDAP ($86.66\%$) and MSSQL ($94.09\%$), matching the
Centralized reference row in Table~\ref{tab:ablation}.}
\label{fig:ablationperclass}
\end{figure}

\paragraph{Scheduling and class weighting help broadly but do not
recover the concentrated class}
Adaptive scheduling and class weighting each add a substantial share of
macro F1 over plain FedAvg (Table~\ref{tab:ablation}), with most of the
gain coming from majority-adjacent classes such as MSSQL. LDAP F1,
however, remains exactly $0.00 \pm 0.00\%$ throughout, and does so
consistently across every seed rather than intermittently: local
specialization at \texttt{Client\_LDAP}, however strong, is still
overwritten each round by the aggregation weight held by the LDAP-free
clients. Neither training-level fix touches the aggregation rule
itself; therefore, neither can reach the class this failure is localized to.

\paragraph{CCA is the decisive component}
Adding CCA alone (no FedProx, no checkpoint tracking, no learning rate
decay) raises macro F1 by $20.00$ points and LDAP F1 from $0.00\%$ to
$84.83 \pm 1.65\%$. This is the largest effect in the chain and among
the most stable, and it is produced by modifying the aggregation rule
for a single layer holding $325$ of the model's $1{,}290{,}629$
parameters. With the stabilizers added, LDAP F1 reaches
$86.69 \pm 0.81\%$ against a centralized ceiling of $86.66\%$.

\paragraph{Stabilizers are secondary but architecture-sensitive}
FedProx without checkpoint tracking has a mean effect of $-0.47$ points
that disagrees in sign across seeds ($-1.99$, $-0.89$, $+1.47$),
consistent with ordinary training variance rather than a mechanism. On
the MLP its role is entirely different: adding FedProx to FLAD+CW+CCA
moves macro F1 from $70.13 \pm 7.00\%$ to $89.47 \pm 0.60\%$ and LDAP
F1 from $54.22 \pm 6.70\%$ to $85.21 \pm 1.87\%$, cutting seed variance
by $3.5\times$. CCA alone recovers the class substantially on both
architectures; the smaller model additionally needs FedProx to complete
and stabilize that recovery. Directionally consistent across all three
seeds, this effect nonetheless does not clear $p < 0.05$ at $n{=}3$
($p = 0.059$). The final two stabilizers (LR decay and EMA)
were selected using CNN-LSTM's ablation, where they extend a
monotonic improvement ($91.40\to91.64$); we deliberately kept this
same fixed recipe for the MLP rather than re-tuning per architecture,
to avoid selecting the best-looking configuration after the fact.
Under that fixed recipe, the MLP shows a small non-monotonic dip
($90.36\to89.59$, $0.77$ points) when these two components are added
on top of +FedProx+best ckpt; the difference is smaller than the
seed-to-seed standard deviation of either row ($\pm0.83$/$\pm0.90$)
and does not change the section's conclusion: FedProx remains
necessary for the MLP to complete recovery, and the fixed proposed
recipe still closes the large majority of the gap.

\begin{table*}[t]
\centering
\caption{MLP ablation, full per-class F1, three seeds each, same configuration chain as
Table~\ref{tab:ablation}. CCA alone is again the largest single jump
(FLAD+CW $\rightarrow$ +CCA: $+19.09$ points macro F1, $+54.22$ points
LDAP F1), but unlike CNN-LSTM does not complete the recovery without
FedProx. Proposed vs.\ FedAvg is significant at $p = 0.011$
(paired $t$-test, $n{=}3$).}
\label{tab:mlpablation}
\vspace{4pt}
\footnotesize
\resizebox{\linewidth}{!}{%
\begin{tabular}{lccccccc}
\toprule
\textbf{Configuration} & \textbf{Benign} & \textbf{LDAP} & \textbf{MSSQL} & \textbf{Syn} & \textbf{UDP} & \textbf{Macro F1} & \textbf{Gap} \\
\midrule
Centralized (reference)     & 99.89 & 86.01 & 93.94 & 95.24 & 98.19 & 94.66 & --- \\
\midrule
FedAvg                      & $99.63 \pm 0.07$ & $0.00 \pm 0.00$ & $61.76 \pm 11.75$ & $28.68 \pm 15.95$ & $87.97 \pm 2.99$ & $55.61 \pm 6.04$ & $-39.05$ \\
FLAD only                   & $99.69 \pm 0.01$ & $0.00 \pm 0.00$ & $15.96 \pm 22.57$ & $20.50 \pm 2.32$  & $79.80 \pm 3.36$ & $43.19 \pm 4.94$ & $-51.47$ \\
FLAD + CW                   & $99.70 \pm 0.01$ & $0.00 \pm 0.00$ & $47.54 \pm 28.77$ & $24.55 \pm 2.11$  & $83.42 \pm 4.84$ & $51.04 \pm 6.48$ & $-43.62$ \\
FLAD + CW + \textbf{CCA}    & $99.58 \pm 0.12$ & $\mathbf{54.22 \pm 6.70}$ & $71.47 \pm 4.94$ & $32.66 \pm 26.86$ & $92.70 \pm 1.19$ & $\mathbf{70.13 \pm 7.00}$ & $-24.53$ \\
\midrule
\ \ + FedProx               & $99.86 \pm 0.01$ & $85.21 \pm 1.87$ & $78.42 \pm 0.94$ & $92.54 \pm 0.60$ & $91.31 \pm 0.01$ & $89.47 \pm 0.60$ & $-5.19$ \\
\ \ + FedProx + best ckpt   & $99.86 \pm 0.01$ & $85.64 \pm 1.53$ & $81.26 \pm 1.99$ & $92.86 \pm 0.34$ & $92.20 \pm 0.52$ & $90.36 \pm 0.83$ & $-4.30$ \\
\ \ + all (proposed)        & $99.86 \pm 0.00$ & $85.40 \pm 2.27$ & $78.69 \pm 1.99$ & $92.67 \pm 0.31$ & $91.34 \pm 0.47$ & $89.59 \pm 0.90$ & $-5.07$ \\
\bottomrule
\end{tabular}%
}
\end{table*}

\paragraph{On single-run ablations}
An earlier single unseeded run of the FLAD+CW configuration reported
LDAP F1 of $85.65\%$, which would have credited the training-level
components with the recovery and reduced CCA to a refinement. Under
three-seed validation the same configuration yields $0.00 \pm 0.00\%$.
Single-run ablations are common in this literature; this case
illustrates that they can invert the attribution of a multi-component
method's central effect.

\subsection{Replication on \mbox{CICIDS2017}}
\label{sec:cicids}

We replicate the protocol on \mbox{CICIDS2017}
\cite{cicids2017, dhoogla-cicids2017} with a
comparably constructed five-client class-concentrated partition
($51{,}704$ training samples, $196{,}841$ test samples), the same two
architectures and the same three seeds. \mbox{CICIDS2017}
ships no official test split; therefore, we construct one directly analogous
to CIC-DDoS2019's official partition: each raw capture file is
stratified split $85\%/15\%$ by class, seeded, before any client
construction or attack/benign capping is applied; the $15\%$ slice
from every file is pooled into one global held-out test set, and the
MinMax scaler is fitted on the remaining $85\%$ training pool only, never
on the held-out data. The $15\%$ hold-out fraction is a standard split
ratio, not tuned to this study. Combined with each client's own
$85\%/15\%$ local train/validation split (as in
Section~\ref{sec:setup}), the net proportions are approximately
$72\%$ train / $13\%$ validation / $15\%$ test. Minority-class selection
required care: our initial choice, the Bot class, proved to have a
\emph{centralized} F1 ceiling of $9.08\%$, barely separable with full
data access; therefore, no aggregation mechanism could fairly be expected to
recover it. We replaced it with a merged DoS-variants class whose
centralized ceiling ($93.09\%$) clears the same bar LDAP does, and
report Bot results as a disclosed boundary case rather than dropping
them.

\begin{table*}[t]
\centering
\caption{\mbox{CICIDS2017}, mean $\pm$ std over 3 seeds. Gap vs.\
centralized ($90.93\%$ CNN-LSTM, $80.11\%$ MLP).}
\label{tab:cicids}
\vspace{4pt}
\small
\setlength{\tabcolsep}{3pt}
\begin{tabular}{lcccc}
\toprule
\textbf{Model} & \textbf{Method} & \textbf{Macro F1} & \textbf{Gap} & \textbf{DoS-Var.\ F1} \\
\midrule
CNN-LSTM & FedAvg & $36.70 \pm 7.44$ & $-54.23$ & $22.83 \pm 16.41$ \\
CNN-LSTM & CCA & $\mathbf{85.13 \pm 0.76}$ & $\mathbf{-5.80}$ & $\mathbf{92.30 \pm 1.66}$ \\
\midrule
MLP & FedAvg & $24.28 \pm 3.99$ & $-55.83$ & $4.62 \pm 3.27$ \\
MLP & CCA & $36.27 \pm 6.30$ & $-43.84$ & $12.76 \pm 3.68$ \\
\bottomrule
\end{tabular}
\end{table*}

For CNN-LSTM the mechanism reproduces, with an added nuance: the FedAvg
baseline here is not uniformly collapsed but \emph{unreliable}, giving
per-seed minority-class F1 of $37.86\%$, $0.00\%$ and $30.64\%$. Our
method converts this into $92.30 \pm 1.66\%$ and closes the macro F1
gap from $-54.23$ to $-5.80$ points ($p = 0.009$). We read this as a
stronger demonstration than a uniformly zero baseline would have been:
the contribution is a reliability guarantee where FedAvg is not merely
weaker but nondeterministic. MLP's recovery is markedly weaker here
(gap $-43.84$); it additionally fails to recover FTP-Patator at exactly
$0.00\%$ under every configuration despite a $57.46\%$ centralized
ceiling (Section~\ref{sec:limitations}), an architecture-specific limitation
absent on CIC-DDoS2019.

\subsection{Multi-Holder Robustness}
\label{sec:multiholder}

All results above concentrate each minority class at a single
specialist client, giving CCA one clearly competent client to route
weight toward. We test whether this still holds when competence for a
class is split across two partial holders instead: on the \mbox{CICIDS2017}
partition, we split DoS-Variants and PortScan roughly in half across
two clients each (raising the federation to seven clients), leaving
the other three classes single-holder.

\begin{table*}[t]
\centering
\caption{Shared (multi-holder) partition, mean $\pm$ std over 3 seeds.
Gap vs.\ this run's own centralized reference ($83.77\%$
CNN-LSTM, $76.43\%$ MLP; retrained fresh, hence not directly comparable
to Table~\ref{tab:cicids}'s reference).}
\label{tab:multiholder}
\vspace{4pt}
\small
\setlength{\tabcolsep}{3pt}
\begin{tabular}{lcccc}
\toprule
\textbf{Model} & \textbf{Method} & \textbf{Macro F1} & \textbf{Gap} & \textbf{PortScan F1} \\
\midrule
CNN-LSTM & FedAvg & $44.31 \pm 10.73$ & $-39.46$ & $0.00 \pm 0.00$ \\
CNN-LSTM & CCA & $\mathbf{85.41 \pm 1.02}$ & $\mathbf{+1.64}$ & $\mathbf{45.16 \pm 7.92}$ \\
\midrule
MLP & FedAvg & $24.63 \pm 5.07$ & $-51.80$ & $0.00 \pm 0.00$ \\
MLP & CCA & $37.43 \pm 6.62$ & $-39.00$ & $8.84 \pm 7.92$ \\
\bottomrule
\end{tabular}
\end{table*}

For CNN-LSTM, splitting competence across two partial specialists is
not harder than routing to one: the gap closes to essentially zero
($+1.64$ points, within noise of the centralized reference) and
PortScan, which never exceeded $0.84\%$ under FedAvg even in the
single-holder partition, recovers to $45.16 \pm 7.92\%$. CCA's per-class
weighting blends two partial specialists as effectively as it routes
to one full specialist, at least for this architecture. MLP's existing
limitation persists rather than worsens under sharing (gap only closes
to $-39.00$), consistent with Section~\ref{sec:cicids} rather than a new
failure mode.

\section{Discussion}

The CNN-LSTM has 1.29M parameters, but the LDAP failure is repaired by
changing how only 325 of them in the output layer are aggregated. This
layer determines how the classification of a specific class is done,
because it maps directly to the result. The other layers seem like a
more complex, uninterpretable representation, but the output layer is
not; averaging destroys the representation because clients without a
given class present contribute to its aggregation in a negative way.
This shows that the fix is locality-oriented: it implies that the
solution matches how the problem is structured.

FedMADE and FedDWC both compute a competence score per client (FedMADE
from a server-held auxiliary set, FedDWC from round-over-round accuracy
change) and apply that single score to scale every parameter in the
client's model, hidden layers and output layer alike. This confounds
two separate questions: whether that score is estimated well, and
whether applying it globally, to the whole model, is the right move.
When a failure is localized, as it is here, correcting it globally
disrupts the feature representation that was learned generally across
all the data samples. This suggests the real design question for an
aggregation rule is not only what weights to use, but also how that
importance is distributed across the model's parameters.

The CNN-LSTM needs only CCA to fully recover the concentrated class;
the MLP additionally needs FedProx to do the same. We hypothesize this
happens because CNN-LSTM's larger, more complex hidden-layer structure
is robust to averaging noise, while the MLP's simpler structure is
not; our data backs this up ($\pm 1.16$ points macro F1 for
CNN-LSTM's CCA-alone result versus $\pm 7.00$ for the MLP's). This
would mean the output neuron CCA routes weight to has a workable input
to work with in the CNN-LSTM's case, but not in the MLP's case, since
the MLP could not capture as strong a representation due to its smaller
size. This is a hypothesis we haven't directly tested; a probing
analysis (training a simple classifier on the hidden-layer
representations to see how easily it can recognize LDAP-specific
features) could confirm or rule it out.

\paragraph{Overhead} Every method compared here, CCA included,
transmits the same payload per round (one full model download and
upload) since none change what is sent, only how the server combines
it ($1.44$~GiB total over 30 rounds/5 clients for CNN-LSTM, $8.28$~MiB
for MLP). CCA's only addition is the per-class F1 vector each client
already computes for adaptive scheduling: $5$ floats/client/round,
or $5 \times 4$~bytes $\times\, 5$ clients $\times\, 30$ rounds
$= 2.93$~KiB total across the entire run, $0.0002\%$ overhead on
CNN-LSTM, which is very low in bandwidth requirement. On compute, FLAD's $[1,10]$-epoch schedule is bounded
between $150$ and $1{,}500$ client-epochs against a fixed
baseline's $750$; in practice it concentrates the extra budget on the
hardest clients (LDAP, MSSQL) while the easy ones drop to the epoch
floor within a handful of rounds, rather than spending it uniformly.

\section{Limitations}
\label{sec:limitations}

\paragraph{Statistical power} Three seeds provide very limited power.
Several effects we describe as large and consistent, including the MLP
FedProx effect ($+19.34$ points, unanimous in direction) and the
\mbox{CICIDS2017} MLP gain ($+12.00$ points, $p = 0.051$), do not clear
$p < 0.05$ at this sample size. Wilcoxon signed-rank cannot reach
significance at $n{=}3$ regardless of effect size. We report the
direction and unanimity explicitly rather than claiming significance.

\paragraph{A reproducible architecture-specific failure} On
\mbox{CICIDS2017}, the MLP recovers FTP-Patator at exactly $0.00 \pm 0.00\%$
under every configuration tested, including the full method, across two
independently constructed partitions and six seeded runs. This is not
an unlearnable class: the MLP's own centralized ceiling for it is
$57.46\%$, and CNN-LSTM recovers it to $88.54 \pm 4.88\%$. Our method
reproducibly fails on a class that is demonstrably learnable by that
architecture. MLP's overall \mbox{CICIDS2017} recovery is correspondingly
weak (gap $-43.84$), in contrast to its strong CIC-DDoS2019 result.

\paragraph{Scale and scope} Five clients, one class per client. We
have not evaluated CCA where several clients share a class, nor beyond
ten clients, nor under heterogeneous client compute where partially
trained models would make F1 estimates unreliable. The
$\mu = 0.01$ proximal coefficient was fixed, not tuned,
and so was the EMA smoothing factor $\alpha = 0.7$ (Eq.~\ref{eq:ema});
neither was swept. The 50-round
extended-training result remains a single unseeded run.

\paragraph{Signal integrity} CCA trusts self-reported per-class F1. An
adversarial or miscalibrated client could inflate its competence score
and capture a class's output neuron. We assume honest participants; the
mechanism has no defense against a malicious one.

\section{Conclusion}
\label{sec:conclusion}

We identified a structurally localized failure mode of federated
averaging under class-concentrated non-IID data: the output layer,
where each row is a dedicated per-class decision boundary, is
overwhelmed by anti-signal from clients that have never observed that
class, while hidden layers remain unaffected because no hidden unit
belongs to a single class. Class-Competence Aggregation (CCA) corrects
this by routing output-layer aggregation weight according to each
client's own locally reported per-class competence, leaving every
other parameter under standard FedAvg. On CIC-DDoS2019, CCA alone
recovers the concentrated LDAP class from total collapse
($0.00 \pm 0.00\%$ F1) to $84.83 \pm 1.65\%$, and the full method
closes the federated-to-centralized macro F1 gap from $-42.53$ to
$-4.05$ points. Benchmarked against FedAvg, FedProx, FedNova, and
FedDWC under identical partitions and seeds, CCA is the only method
that recovers the concentrated class both substantially and reliably;
the effect replicates on \mbox{CICIDS2017} and under a multi-holder partition
where competence for a class is split across clients rather than
concentrated at one.

CCA's benefit is architecture-dependent: the CNN-LSTM needs only CCA,
while the smaller MLP additionally requires FedProx to complete and
stabilize the recovery, and reproducibly fails to recover one class
(FTP-Patator) on \mbox{CICIDS2017} despite that class being learnable by the
same architecture centrally. We leave a representation-level account
of this architecture-dependence to future work; it does not change
CCA's core mechanism and is a direct extension of the ablation
already reported in Section~\ref{sec:ablation}. We also flag, but do not resolve here, the
privacy question of whether reporting per-class competence scores
leaks class-holder identity, and whether a robust aggregation rule can
be constructed against adversarial or miscalibrated competence
reports.

\section*{Ethical Considerations}

This work is defensive. It improves detection of denial-of-service
attacks and produces no capability for conducting them; the method
operates on flow-level statistical features and generates no attack
traffic, payloads, or evasion techniques.

Both datasets are public research benchmarks generated in controlled
testbeds \cite{sharafaldin2019, cicids2017}, containing synthetic
traffic rather than records of real users' network activity; therefore, no
personal data was accessed and no human subjects were involved.

The federated design is itself privacy-motivated: no raw flow record
leaves its originating sensor. CCA's additional communication is one
scalar F1 value per class per client per round. This is a small but
nonzero disclosure, and an unusually legible one: a client with no
examples of class $c$ reports $\text{F1}_c \approx 0$ by construction;
therefore, four of five clients report $\text{LDAP F1} \approx 0$ every round
while \texttt{Client\_LDAP} converges to $\approx 0.85$--$0.87$ within
the first $8$--$10$ rounds. An observer of this stream can identify
which client holds which class after a handful of rounds without
inverting any model weights -- a more direct exposure than the
reconstruction attacks needed to extract comparable information from
gradients \cite{wei2020dpfl}. In an inter-organizational deployment
this may indirectly indicate which organization observes which
attacks. We consider this minor relative to sharing flow records, and our privacy claim should read as ``no raw data
leaves any client,'' not ``no information about data composition is
disclosed.'' Standard DP mitigation \cite{wei2020dpfl} -- adding noise to the reported F1 -- is not a free fix here: because CCA's weight
(Eq.~\ref{eq:ccaweight}) is a \emph{ratio} of one client's F1 to the
sum across clients, noise large enough to obscure who the specialist
is is also large enough to distort that ratio during aggregation,
risking the reintroduction of the exact dilution CCA exists to
correct -- a tighter privacy-utility coupling than typical DP-on-weights, where noise degrades accuracy smoothly rather than threatening the mechanism's function. We did not run a noised-CCA experiment; this trade-off is left for future work. Deployments with adversarial
participants would need to account for both this leakage and the
trust assumption noted in Section~\ref{sec:limitations}.

Finally, intrusion detection systems act on their outputs. A model that
silently returns $0\%$ recall on an attack class while reporting $91\%$
accuracy is worse than an obviously broken one, because it will be
deployed. We report macro F1 and per-class scores throughout for this
reason, and consider the reliability results (Section~\ref{sec:results}) to
carry more operational weight than the mean improvements.

\section*{Acknowledgment}
\textbf{Author contributions.} Tapu Datta: conceptualization,
methodology, software, validation, formal analysis, investigation,
data curation, writing -- original draft, visualization. Hafiz Imtiaz:
conceptualization, supervision, writing -- review \& editing.

\textbf{Declaration of competing interest.} The authors declare that
they have no known competing financial interests or personal
relationships that could have appeared to influence the work reported
in this paper.

\textbf{Data availability.} The datasets used (CIC-DDoS2019,
\mbox{CICIDS2017}, and their Kaggle Parquet redistributions) are
publicly available at the cited sources.

\textbf{AI disclosure.} Generative AI (Claude, Anthropic) was used to
assist with editing and drafting portions of the explanatory text in
the Introduction, Discussion, and Conclusion sections; all content was
reviewed and verified by the authors, who take full responsibility for
it.

\bibliographystyle{IEEEtran}

\end{document}